\documentclass[letterpaper]{article} 
\usepackage{aaai2026}  
\usepackage{times}  
\usepackage{helvet}  
\usepackage{courier}  
\usepackage[hyphens]{url}  
\usepackage{graphicx} 
\usepackage{natbib}  
\usepackage{caption} 
\usepackage{algorithm}
\usepackage{algorithmic}
\usepackage{utfsym}
\usepackage{amssymb}
\usepackage{makecell}
\usepackage{booktabs}
\usepackage{multirow}
\usepackage{xcolor}
\usepackage{amsmath} 
\usepackage{newfloat}
\usepackage{listings}
\DeclareCaptionStyle{ruled}{labelfont=normalfont,labelsep=colon,strut=off} 
\floatstyle{ruled}
\newfloat{listing}{tb}{lst}{}
\floatname{listing}{Listing}
\title{Training-Free Anomaly Generation via Dual-Attention Enhancement in Diffusion Model}
\author{
    Zuo Zuo\textsuperscript{\rm 1,2},
    Jiahao Dong\textsuperscript{\rm 2},
     Yanyun Qu\textsuperscript{\rm 3},
      Zongze Wu\textsuperscript{\rm 1,2,4} \thanks{Corresponding author.}\\
}
\affiliations{
    \textsuperscript{\rm 1}National Key Laboratory of Human-Machine Hybrid Augmented Intelligence, and Institute of Artificial
Intelligence and Robotics, Xi’an Jiaotong University\\

    \textsuperscript{\rm 2}Guangdong Laboratory of Artificial Intelligence and Digital Economy (SZ) \\ 
    \textsuperscript{\rm 3}Xiamen University\\ 
    \textsuperscript{\rm 4}College of Mechatronics and Control Engineering, Shenzhen University \\
    Nostalgiaz@stu.xjtu.edu.cn
}

\usepackage{bibentry}

\begin{document}

\maketitle

\begin{abstract}
Industrial anomaly detection (AD) plays a significant role in manufacturing where a long-standing challenge is data scarcity. A growing body of works have emerged to address insufficient anomaly data via anomaly generation. However, these anomaly generation methods suffer from lack of fidelity or need to be trained with extra data. To this end, we propose a training-free anomaly generation framework dubbed AAG, which is based on Stable Diffusion (SD)’s strong generation ability for effective anomaly image generation. Given a normal image, mask and a simple text prompt, AAG can generate realistic and natural anomalies in the specific regions and simultaneously keep contents in other regions unchanged. In particular, we propose Cross-Attention Enhancement (CAE) to re-engineer the cross-attention mechanism within Stable Diffusion based on the given mask. CAE increases the similarity between visual tokens in specific regions and text embeddings, which guides these generated visual tokens in accordance with the text description. Besides, generated anomalies need to be more natural and plausible with object in given image. We propose Self-Attention Enhancement (SAE) which improves similarity between each normal visual token and anomaly visual tokens. SAE ensures that generated anomalies are coherent with original pattern. Extensive experiments on MVTec AD and VisA datasets demonstrate effectiveness of AAG in anomaly generation and its utility. Furthermore, anomaly images generated by AAG can bolster performance of various downstream anomaly inspection tasks.
\end{abstract}

%

\section{Introduction}
\begin{figure}[t]
  \centering
  \centerline{\includegraphics[width=0.9\linewidth]{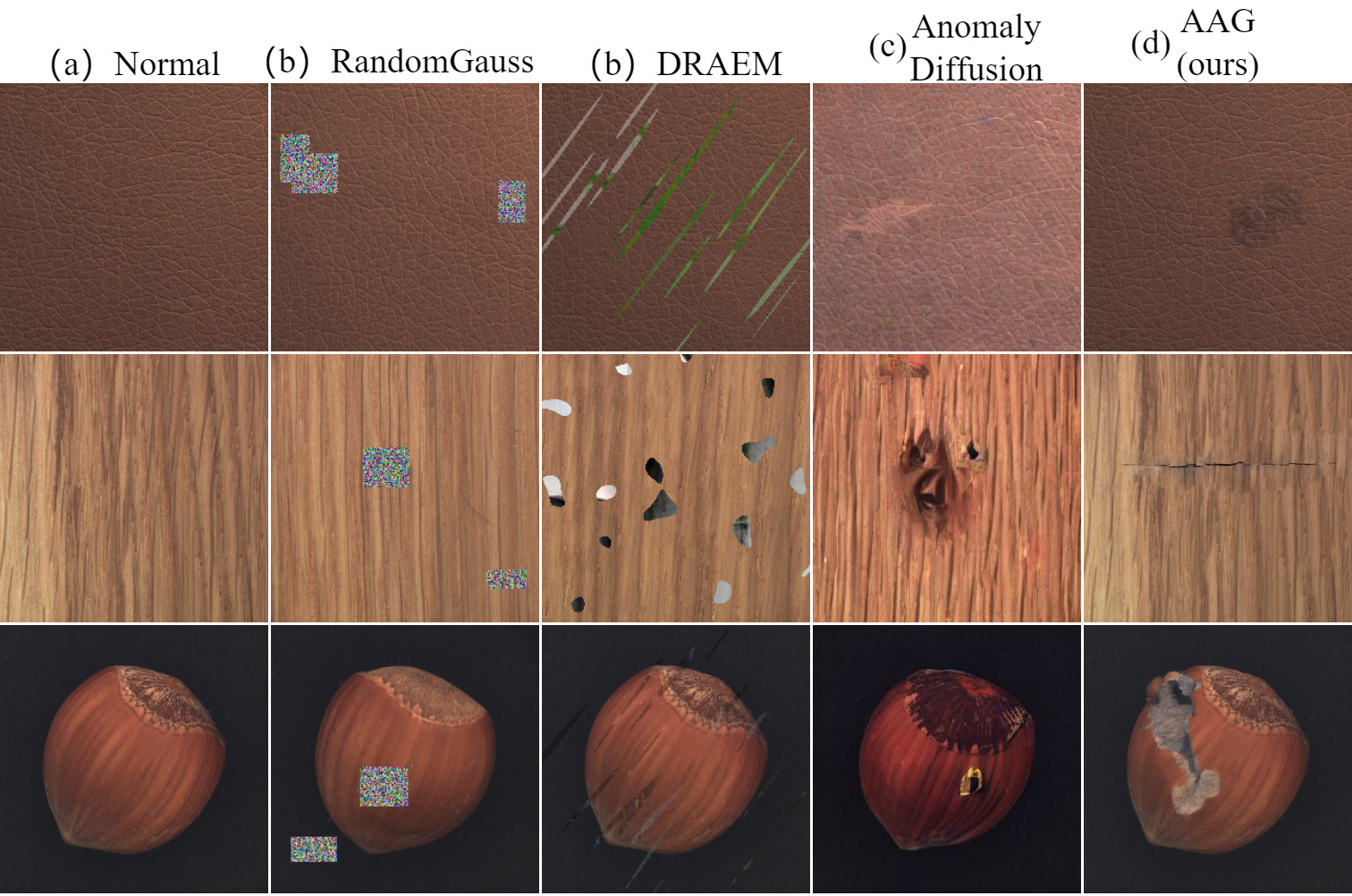}}
  \caption{Examples of synthetic anomalies generated by different methods.}
  \label{fig::intro_compare}
\end{figure}

Industrial anomaly detection (IAD) has gained increasing attention recently which aims to detect anomalies in given samples~\cite{glass}. IAD is an indispensable task in industrial scenarios, ensuring product quality and safety. But insufficient anomaly data poses a dilemma to IAD and data collecting is difficult in industrial scenarios~\cite{clipfsac}. Moreover, constructing large multi-modal industrial models is data-hungry~\cite{anomalyov} and data scarcity dramatically hinders the development of large multi-modal models in industrial scenarios. To solve lack of anomaly samples, most unsupervised anomaly detection methods generate anomaly samples to meet data demands. Anomaly generation intends to generate anomalies on specific regions and keep other regions unchanged. Natural synthetic anomalies (NSA)~\cite{nsa} integrates Poisson image editing and Gamma distribution-based patch shape sampling strategy in generation process. In CDO~\cite{cdo}, RandomGauss randomly selects some square regions and replace them with random values sampled from a Gaussian normal distribution. DRAEM~\cite{draem} samples anomaly patterns from image set and places these patterns on the anomaly free image according to mask. However, anomaly images generated by these traditional methods are unreal and inconsistent in Figure \ref{fig::intro_compare}. 

Recent academic attention has been geared toward anomaly generation based on generative models. DFMGAN~\cite{dfmgan} trains StyleGAN2 as backbone and incorporates defect-aware residual blocks into backbone for generating defect images. But it involves a large amount of training samples which is unexpected in industrial scenarios. Anomalydiffusion~\cite{anomalydiffusion} leverages the strong prior information of a pretrained LDM and uses few-shot anomaly images to optimize anomaly embedding and spatial encoder. However, type and data of each defect are necessary to be reached before training, which is very difficult in industrial scenarios. Additionally, quality and authenticity of anomaly images generated by Anomalydiffusion are not guaranteed as illustrated Figure \ref{fig::intro_compare}. Like Anomalydiffusion, AnomalyControl~\cite{anomalycontrol} also uses anomaly data in testing dataset for training and need anomaly caption which means that types of detects must be obtained before training.

\begin{table}[t]
\caption{Comparisons with previous anomaly generation methods in recipe of framework. }
\label{paucbtad}
\centering
\resizebox{0.98\linewidth}{!}{
\begin{tabular}{c|cccc}
\hline
Method&Training &Anomaly Types & Anomaly Samples & Mask \\ \hline
AnomalyXFusion \cite{anomalyxfusion} & \usym{2714} &\usym{2714}  &  \usym{2718}
 &\usym{2714}\\
AnomalyDiffusion \cite{anomalydiffusion} & \usym{2714} &\usym{2718} &\usym{2714} & \usym{2714}\\
AnomalyControl  \cite{anomalycontrol} & \usym{2714} &\usym{2714} & \usym{2714} & \usym{2714}\\
AAG (ours) \cite{padim} & \textcolor{red}{\usym{2718}} &\textcolor{red}{\usym{2718}} &\textcolor{red}{\usym{2718}}& \textcolor{green}{\usym{2714}}\\

\hline
\end{tabular}}
\label{tab::intro::compare}
\end{table}

To holistically address the issues mentioned above, we propose a simple yet effective training-free anomaly generation framework named AAG without training and types of defects, which is based on Stable Diffusion (SD). To generate anomaly images, we set text prompt template to "A [cls] that is damaged and broken" where cls is object type in each dataset. Text prompt is encoded by CLIP text encoder to steer the denoising diffusion process. To enhance the guidance of text prompts and generate pronounced anomalies, we propose Cross-Attention Enhancement (CAE) to re-engineers the cross-attention mechanism within Stable Diffusion. Specifically, we enlarges the similarity between visual tokens in specific regions where we generate anomalies and text tokens, which further enhances the interaction between anomaly visual tokens in and text embedding. To obviate that generated anomalies are not coherent with original content, we propose Self-Attention Enhancement (SAE) which enhances the relationship between anomaly visual tokens and other tokens. We re-weight the similarity matrix in SD's self-attention modules and improve contributions of normal visual tokens when anomaly visual tokens are generated. AAG significantly simplifies anomaly generation framework compared with other anomaly generation methods as illustrated in Table \ref{tab::intro::compare}. We conduct extensive experiments to demonstrate that AAG can generate high-quality anomaly images and dramatically improve performance of downstream anomaly inspection tasks.
The main contribution of this paper can be summarized as follows:
\begin{itemize}
\item We propose a training-free and anomaly-agnostic anomaly generation framework named AAG. AAG gets rid of anomaly training samples and types of anomalies, which significantly simplifies anomaly generation framework and is convenient to deploy. 

\item We design Cross-Attention Enhancement (CAE) and Self-Attention Enhancement (SAE) in which we re-engineer cross-attention and self-attention mechanism in Stable Diffusion. CAE and SAE ensure that we can generate pronounced and coherent anomalies for given images.

\item Experimental results show superior quality and fidelity of anomalies generated by AAG. We use generated anomaly images in different down-stream anomaly inspection tasks, which further demonstrates effectiveness and utility of AAG.
\end{itemize}

 \begin{figure*}[t]
  \centering
  \centerline{\includegraphics[width=0.9\linewidth]{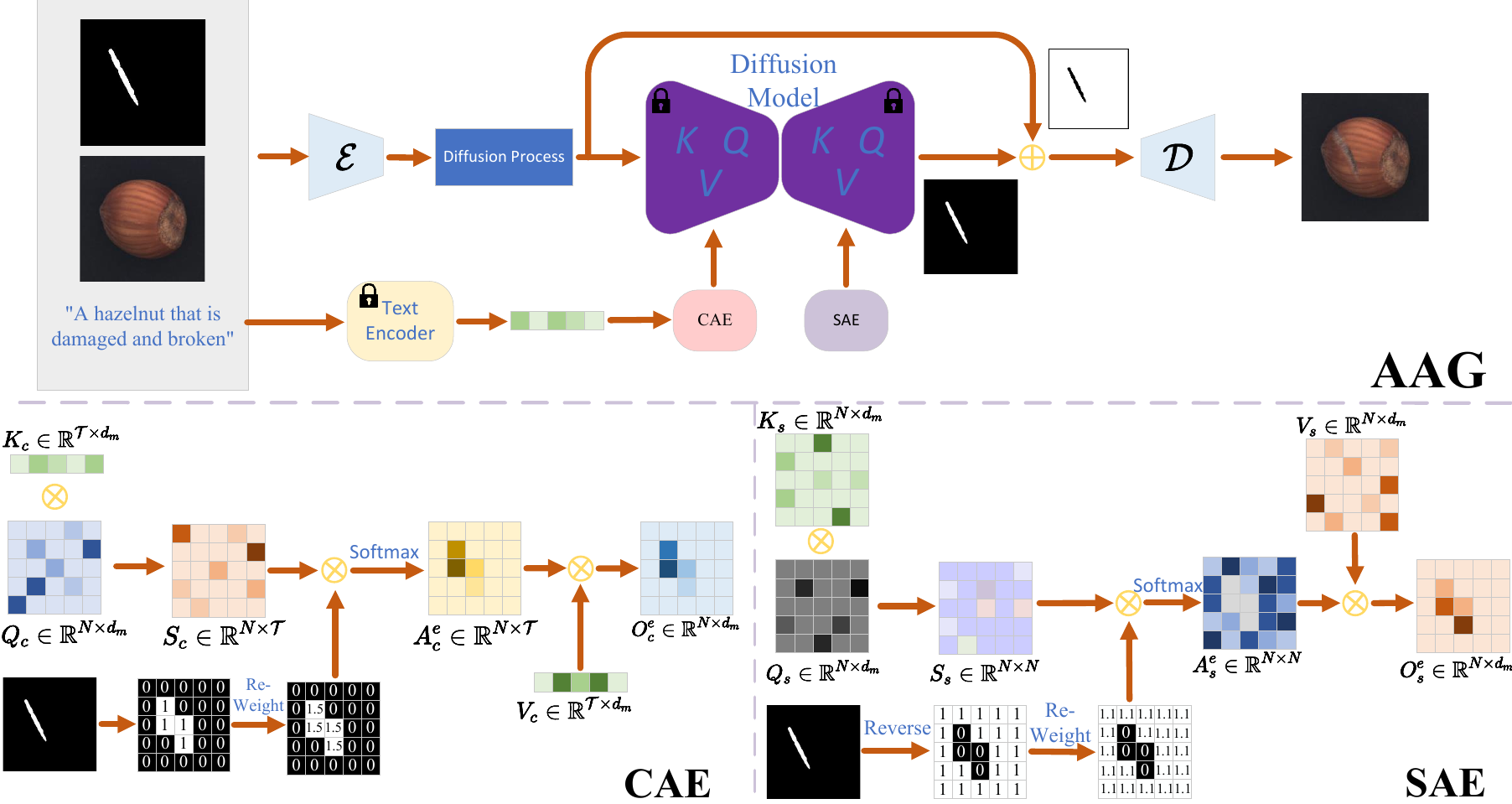}}
  \caption{Overview of the proposed AAG framework.}
  \label{fig::framework}
\end{figure*}

\section{Related Work}
\subsection{Diffusion Models}
The diffusion model is proposed by DDPM~\cite{ddpm} inspired by the principles of nonequilibrium thermodynamics~\cite{eqt}. In the forward process of DDPM, input image is converted into random noise by gradually adding Gaussian noise. Conversely, random noise is removed gradually as a deterministic generation procedure. By employing a pretrained VAE, the latent diffusion model (LDM)~\cite{ldm} operates generation in the latent space, significantly cutting down on both resource consumption and training/inference time. To control the generation process, lots of works delve into guidance techniques~\cite{sag, pag, pap}. Classifier guidance~\cite{classifierguidance} in diffusion models uses gradients from a pretrained classifier during sampling to steer the generation toward a target class or attribute, enhancing control without retraining the model. However, classifier-free~\cite{clsfree} guidance  achieves conditional generation without an external classifier by jointly training the model to handle both conditional and unconditional denoising, then blending their outputs during sampling for controlled generation. ControlNet~\cite{controlnet} supports lot of guidance such as text guidance, mask guidance and so on. Our proposed AAG uses simple text prompt to guide diffusion model to generate anomaly images effectively.

\subsection{Anomaly Generation}
Significant efforts have been dedicated to alleviate data scarcity in industrial scenarios. One of the most widely used approaches is anomaly generation. Anomaly generation methods can be categorized into two types: cut-and-paste methods and generative model-based methods. Cut-and-paste methods usually cut partial pixels from source images and combine these pixels with given normal images~\cite{cutpaste, nsa, draem}. These methods are simple but they lack of authenticity and diversity shown in Figure \ref{fig::intro_compare}. Anomaly images generated by generative model-based methods such as generative adversarial networks (GANs) and diffusion models are more realistic~\cite{dfmgan, defgan, anomalycontrol}. DFMGAN~\cite{dfmgan} is based on GAN and generate realistic defect images associated with defect masks, via feature manipulation using defect-aware residual blocks. AnomalyDiffusion~\cite{anomalydiffusion} based on diffusion model disentangles anomalies into anomaly embedding and spatial embedding and adaptive attention re-weighting mechanism to generate anomalies. However, these generative model-based methods need anomaly testing samples to train or availability to kinds of anomalies, which hinders utility and deployment. Our AAG without training and captions achieves anomaly generation cost-efficiently.

\section{Preliminaries}
\textbf{Diffusion Models.} Diffusion models are inspired by non-equilibrium thermodynamics, which consist of forward process, also known as diffusion process, and reverse diffusion process. For an image $x_0$, forward process is defined as:
 \begin{equation}
\begin{aligned}
    x_{t}=x_{0} \sqrt{\bar{\alpha}_{t}}+\epsilon_{t} \sqrt{1-\bar{\alpha}_{t}}, \quad \epsilon_{t} \sim \mathcal{N}(\mathbf{0}, \mathbf{I}),
\end{aligned}
\label{eq:xt}
\end{equation}
where $t$ is randomly sampled from $\{0, 1, ... , T\}$. Forward process is Gaussian noise addition process according to a variance schedule $\beta_1, \dots ,\beta_T$ and $\bar{\alpha}_{t}=\prod_{i=0}^{t} \alpha_{i}=\prod_{i=0}^{t}(1-\beta_{i})$ and $\beta_{i} \in(0,1)$. While reverse diffusion process is step-by-step noise removing process, which is defined as:
\begin{equation}
\begin{aligned}
x_{t-1}=\frac{1}{\sqrt{\alpha_{t}}}\left(x_{t}-\frac{1-\alpha_{t}}{\sqrt{1-\bar{\alpha}_{t}}} \epsilon_{\theta}\left(x_{t}, t\right)\right)+\sigma_{t} z,
\end{aligned}
\label{eq:xt-1}
\end{equation}
where $z \sim \mathcal{N}(\mathbf{0}, \mathbf{I})$. 

The optimization objective of diffusion models is:
\begin{equation}
\begin{aligned}
\mathcal{L}=\mathbb{E}_{t \sim[1-T], x_{0} \sim q\left(x_{0}\right), \epsilon \sim \mathcal{N}(0, \mathbf{I})}\left[\left\|\epsilon-\epsilon_{\theta}\left(x_{t}, t\right)\right\|^{2}\right].
\end{aligned}
\label{eq:diffusion loss}
\end{equation}
where $\epsilon_{\theta}\left(x_{t}, t\right)$ represents the learnable U-Net-like architectures.

\textbf{Attention in Stable Diffusion.} Attention module is one of utmost importance module in Stable Diffusion. In Stable Diffusion, there exist cross-attention module and self-attention module. Let $\mathcal{F}_t\in \mathbb{R}^{h\times w \times c}$ present any latent feature map. For self-attention mechanism, $\ell_{Q}$ , $\ell_{K}$ and $\ell_{V}$ presents learned query key and value linear layers.The self-attention module can be defined as follows:
\begin{equation}  
Q_{self} = \ell_{Q}\left ( \mathcal{F}_{t} \right ), K_{self} = \ell_{K}\left ( \mathcal{F}_{t} \right ), 
\end{equation}  
\begin{equation}  
S_{self} = Q_{self}\left ( K_{self} \right )^{T} /\sqrt{d_{self}},
\end{equation}  
\begin{equation}  
A_{self} = \text{softmax}\left ( S_{self}\right ),
\end{equation}  
where $d_{self}$ is the dimension of the keys and queries. $S_{self}\in \mathbb{R}^{\left ( h\times w \right ) \times \left ( h\times w \right )}$ is similarity matrix and $A_{self}\in \mathbb{R}^{\left ( h\times w \right ) \times \left ( h\times w \right )}$ is self-attention map which can control the spatial layout and shape details of the generated image~\cite{tucs}. For cross-attention mechanism, textual embedding $\mathcal{P}_{e}\in \mathbb{R}^{\mathcal{T}\times d_t}$ is used to guide generation process.  Cross-attention mechanism is defined as follows:
\begin{equation}
 Q_{cross} = \ell_{q}(\mathcal{F}_{t}), \quad K_{cross} = \ell_{k}(\mathcal{P}_{e})
\label{equation3}
\end{equation}
\begin{equation}  
S_{cross} = Q_{cross}\left ( K_{cross} \right )^{T} /\sqrt{d_{cross}},
\end{equation}  
\begin{equation}
 A_{cross} = \text{Softmax}\left(S_{cross}\right)
\label{equation4}
\end{equation}
where $d_{cross}$ is the dimension of the keys and queries. $S_{cross}\in \mathbb{R}^{\left ( h\times w \right ) \times N}$ is vision-text similarity matrix and $A_{cross}\in \mathbb{R}^{\left ( h\times w \right ) \times \mathcal{T}}$ is cross-attention map. Cross-attention map can steer generation to follow the prompt.

\section{Method}
The overall pipeline of AAG is illustrated in Figure \ref{fig::framework}. Without training, AAG can be utilized to generate authentic anomalies on given samples cost-efficiently based on Stable Diffusion. The inputs of AAG are normal image $i_n$, mask $M$ and a text prompt with fixed template $\mathcal{P}$. Mask $M$ controls where we want generate anomalies. The core mechanism in AAG is cross-attention enhancement (CAE) and self-attention enhancement (SAE). CAE enhances the similarity between visual representations in selected regions and text embedding while inhibiting the influence of text embedding on other regions, which improve alignment between the generated anomalies and mask $M$.  CAE takes effect in cross-attention module in Stable Diffusion. SAE directly manipulates the similarity matrix in cross-attention module of Stable Diffusion, which magnifies the similarity between regions within mask and other regions. This implies that the influence of regions outside of mask increases when anomaly regions are generated. Re-engineering the self-attention mechanism in SAE will make generated anomalies more seamless and avoid generating abrupt anomalies. Additionally, to be spared from changing contents outside of mask in the generation process, we utilize blended mechanism with mask $M$ to maintain these contents unchanged.

 \begin{figure}[t]
  \centering
  \centerline{\includegraphics[width=0.84\linewidth, ]{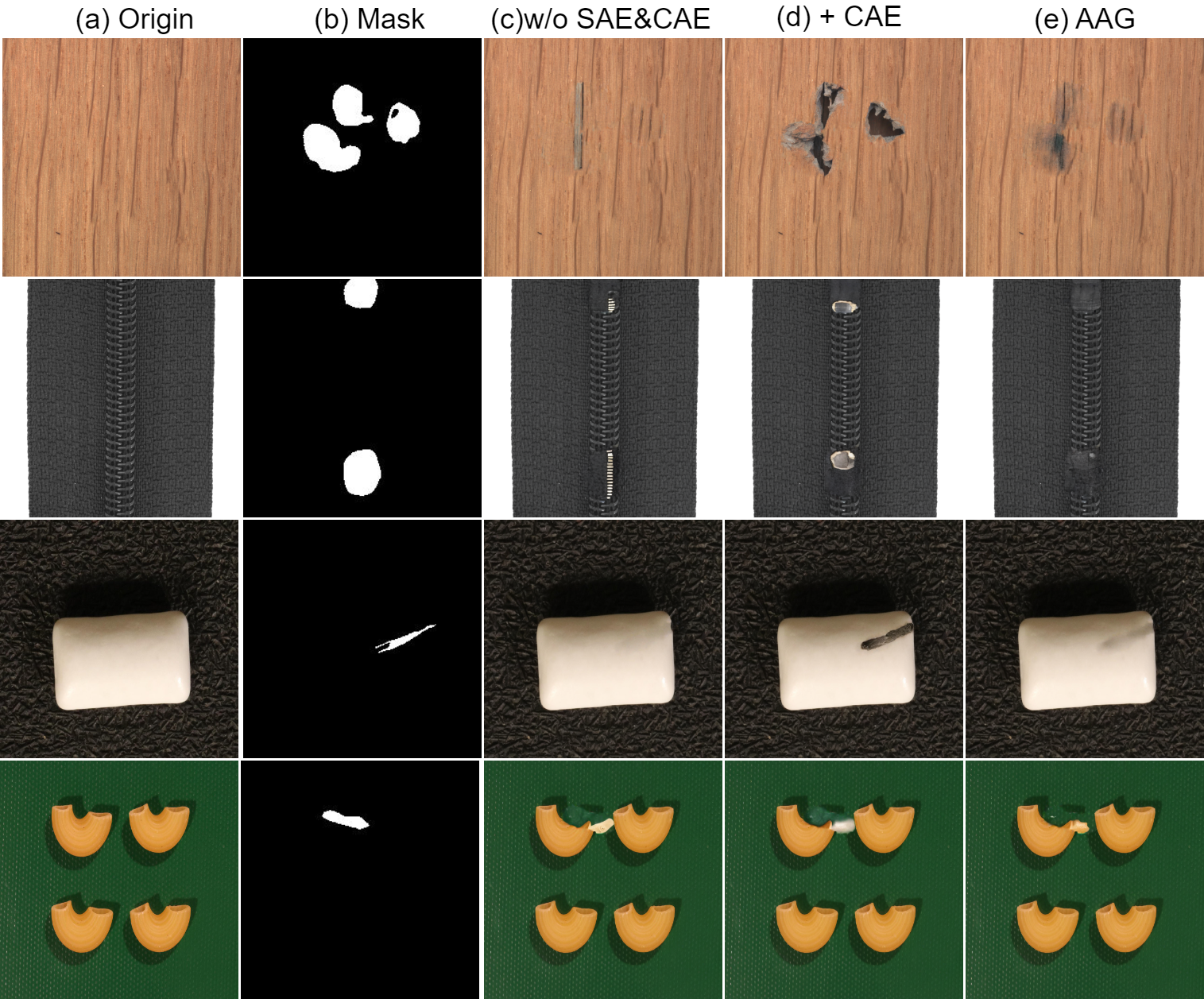}}
  \caption{Comparison between generation results with and without SAE and CAE modules.}
  \label{fig::method_saecae}
\end{figure}

\subsection{Cross-Attention Enhancement}
In cross-attention module of Stable Diffusion, the interaction between the visual tokens and the text embedding is established and guides generation process. Let specific feature map of $l$ cross-attention layer in the U-net~\cite{unet} be $\mathcal{F}^l\in \mathbb{R}^{N\times d}$ where $N=h\times w$. The text embedding is denoted as $\mathcal{P}_{e}\in \mathbb{R}^{\mathcal{T}\times d_t}$. The cross-attention mechanism is defined as:
\begin{equation}
 Q_{c} = \ell_{q}(\mathcal{F}^l), \; Q_{c}\in \mathbb{R}^{N\times d_m}, 
\label{equation3}
\end{equation}
\begin{equation}
 K_{c} = \ell_{k}(\mathcal{P}_{e}),V_{c} = \ell_{v}(\mathcal{P}_{e}), \; K_{c},V_{c}\in \mathbb{R}^{\mathcal{T}\times d_m},
\label{equation3}
\end{equation}
\begin{equation}  
S_{c} = Q_{cross}\left ( K_{cross} \right )^{T} /\sqrt{d_{cross}}, \; S_{c}\in \mathbb{R}^{N\times \mathcal{T}},
\end{equation}  
\begin{equation}
 A_{c} = \text{Softmax}\left(S_{c}\right), \;  A_{c}\in \mathbb{R}^{N\times \mathcal{T}}.
\label{equation4}
\end{equation}
where $d_{m}$ is the dimension of the keys and queries. $S_{c}$ is vision-text similarity matrix and $A_{c}$ is cross-attention map. The output feature map $O_c$ is defined as $O_c=A_{c}V_{c},O_c\in \mathbb{R}^{N\times d_m}$. $A_{c}^{i,j}$ represents the weight of text token $j\mbox{-}th$ when visual token $i$ is generated~\cite{p2p,ae}. That is,  the value in the $i\mbox{-}th$ row of the attention map $A_{c}$ indicates the degree to which every text token in $V_{c}$ contributes to the generation of visual token $O_c^i$. For anomaly generation, we hope that realistic and faithful anomalies are generated on specific regions based on given mask $M$. The mask $M$ is $0\mbox{-}1$ matrix defined as:
\begin{equation}
M_{i,j}=\left\{\begin{array}{ll}
1, & if \; anomaly\; generation \; at \; location \; (i,j) \\
0, & otherwise.
\end{array}\right.
\end{equation}

\begin{algorithm}[tb]
\caption{The Pipeline of Anomaly Generation in AAG}
\label{alg:algorithm}
\textbf{Input}: Normal image $x$, text prompt $\mathcal{P}$, timestep $T$, input mask $M$, U-net model with SAE and CAE $\epsilon_{\theta}\left(x_{t}, \mathcal{P}_e, t\right)$ \\
\textbf{Output}: Generated anomaly image $x_n$ based on normal image $x$
\begin{algorithmic}[1] 
\STATE $z_0^{\prime} = VAE\_Encoder(x)$
\STATE $z_{t}=z_{0}^{\prime} \sqrt{\bar{\alpha}_{t}}+\epsilon_{t} \sqrt{1-\bar{\alpha}_{t}}, \quad \epsilon_{t} \sim \mathcal{N}(\mathbf{0}, \mathbf{I})$
\STATE $\mathcal{P}_e = Text\_encoder(\mathcal{P})$
\FOR{all $t$ from $T$ to 0}
\STATE $z_{t-1}^{init}=\frac{1}{\sqrt{\alpha_{t}}}\left(z_{t}-\frac{1-\alpha_{t}}{\sqrt{1-\bar{\alpha}_{t}}} \epsilon_{\theta}\left(x_{t}, \mathcal{P}_e, t\right)\right)+\sigma_{t} n,n \sim \mathcal{N}(\mathbf{0}, \mathbf{I})$
\STATE $z_{t}^0=z_{0} \sqrt{\bar{\alpha}_{t}}+\epsilon_{t} \sqrt{1-\bar{\alpha}_{t}}, \quad \epsilon_{t} \sim \mathcal{N}(\mathbf{0}, \mathbf{I}),$
\STATE $z_{t-1} = z_{t-1}^{init} \cdot M + z_{t}^0  \cdot (1-M) $
\ENDFOR
\STATE $x_n=VAE\_Decoder(z_0)$
\STATE \textbf{return} $x_n$
\end{algorithmic}
\end{algorithm}

\begin{table*}[!t]
\setlength{\abovecaptionskip}{0cm}
\centering
\caption{
Generation quantitative results with IS and IC-LPIPS (abbreviated as IL) on MVTec AD dataset. Red indicates the best results, and blue indicates the second-best results. }
\setlength{\tabcolsep}{2pt} 
\resizebox{0.95\linewidth}{!}{
\begin{tabular}{c c|ccccccccccccccc|c}
\hline
Method & Metric & bottle & cable & caps & carp & grid & hazel & leath & metal & pill & screw & tile & brush & trans & wood & zipper & Mean \\
\hline
\multirow{2}{*}{CDC \cite{cdc}}
& IS & 1.52 & 1.97 & 1.37 & \textcolor{blue}{1.25} & 1.97 & 1.97 & 1.80 & 1.55 & 1.56 & 1.13 & 2.10 & 1.63 & 1.61 & 2.05 & 1.30 & 1.65 \\
 & IL & 0.04 & 0.19 & 0.06 & 0.03 & 0.07 & 0.05 & 0.07 & 0.04 & 0.06 & 0.11 & 0.12 & 0.06 & 0.13 & 0.03 & 0.05 & 0.07 \\

\hline
\multirow{2}{*}{DefGAN \cite{defgan}}
& IS & 1.39 & 1.70 & 1.59 & 1.24 & 2.01 & 1.87 & \textcolor{red}{2.12} & 1.47 & 1.61 & 1.19 & 2.35 & \textcolor{blue}{1.85} & 1.47 & 2.19 & 1.25 & 1.69 \\
& IL & 0.07 & 0.22 & 0.04 & 0.12 & 0.12 & 0.19 & 0.14 & 0.30 & 0.10 & 0.12 & 0.22 & 0.03 & 0.13 & 0.29 & 0.10 & 0.15 \\
\hline
\multirow{2}{*}{DFMGAN \cite{dfmgan}}
& IS & 1.62 & 1.96 & 1.59 & 1.23 & 1.97 & 1.93 & \textcolor{blue}{2.06} & 1.49 & 1.63 & 1.12 & 2.39 & 1.82 & 1.64 & 2.12 & 1.29 & 1.72 \\
 & IL & 0.12 & 0.25 & 0.11 & 0.13 & 0.13 & 0.24 & 0.17 & 0.32 & 0.16 & 0.14 & 0.22 & 0.18 & 0.25 & 0.35 & \textcolor{red}{0.27} & 0.20 \\
\hline
\multirow{2}{*}{AnoDiff  \cite{anomalydiffusion}}
& IS & 1.58 & 2.13 & 1.59 & 1.16 & 2.04 & \textcolor{blue}{2.13} & 1.94 & \textcolor{blue}{1.96} & 1.61 & 1.28 & \textcolor{blue}{2.54} & 1.68 & 1.57 & \textcolor{blue}{2.33} & 1.39 & 1.80 \\

& IL & \textcolor{red}{0.19} & 0.41 & \textcolor{blue}{0.21} & 0.24 & 0.44 & 0.31 & 0.41 & 0.30 & 0.26 & \textcolor{blue}{0.30} & \textcolor{red}{0.55} & 0.21 & 0.34 & 0.37 & 0.25 & 0.32 \\
\hline

\multirow{2}{*}{AnoXFusion  \cite{anomalyxfusion}}
& IS & \textcolor{blue}{1.62} & 2.13 & 1.68 & 1.16 
& 2.12 & \textcolor{red}{2.24} & 2.00 & \textcolor{blue}{1.96} & 1.63 & \textcolor{blue}{1.29} & \textcolor{red}{2.55} & 1.80 & 1.64 & 2.09 & 1.37 & 1.82 \\
           
& IL & \textcolor{red}{0.19}  & 0.42& 0.20 & 0.22
& 0.45 & 0.33 & \textcolor{blue}{0.42} & \textcolor{blue}{0.34} &\textcolor{blue}{0.27} & 0.29 &0.52
& 0.22 & \textcolor{blue}{0.36} & \textcolor{blue}{0.38} & \textcolor{blue}{0.26} &\textcolor{blue}{0.33} \\

\hline

\multirow{2}{*}{AnoControl  \cite{anomalycontrol}} 
& IS 
& \textcolor{red}{1.63} & \textcolor{blue}{2.14} & \textcolor{blue}{1.69} & 1.18 & \textcolor{blue}{2.26} & 2.12 & 2.08 & \textcolor{red}{1.98} & \textcolor{blue}{1.64} & 1.25 & \textcolor{blue}{2.54} & 1.80 & \textcolor{blue}{1.66} & 2.20 & \textcolor{blue}{1.40} & \textcolor{blue}{1.84} \\

& IL & \textcolor{red}{0.19} & \textcolor{blue}{0.44} & 0.20 & \textcolor{blue}{0.28} & \textcolor{blue}{0.47} & \textcolor{blue}{0.35} & \textcolor{red}{0.43} & \textcolor{blue}{0.34} & \textcolor{blue}{0.27} & 0.28 & \textcolor{blue}{0.53} & \textcolor{blue}{0.24} & \textcolor{red}{0.39} & \textcolor{red}{0.40} & \textcolor{blue}{0.26} & \textcolor{red}{0.35}\\
\hline
\multirow{2}{*}{AAG (Ours)} 
& IS 
& 1.54 & \textcolor{red}{2.33} & \textcolor{red}{1.87} & \textcolor{red}{1.70} & \textcolor{red}{2.44} & 1.98 & 1.79 & 1.94 & \textcolor{red}{1.87} & \textcolor{red}{1.30} & 1.56 & \textcolor{red}{1.97} & \textcolor{red}{2.00} & \textcolor{red}{2.51} & \textcolor{red}{2.41} & \textcolor{red}{1.95} \\

& IL & \textcolor{blue}{0.17} & \textcolor{red}{0.49} & \textcolor{red}{0.27} & \textcolor{red}{0.31} & \textcolor{red}{0.48} & \textcolor{red}{0.37} & 0.33 & \textcolor{red}{0.39} & \textcolor{red}{0.30} & \textcolor{red}{0.38} & 0.50 & \textcolor{red}{0.26} & \textcolor{blue}{0.36} & \textcolor{blue}{0.38} & \textcolor{red}{0.27} & \textcolor{red}{0.35}\\
\hline
\end{tabular}}
\label{tab::isil}
\end{table*}

If original attention maps $A_{c}$ involve generation process of all visual tokens, some problems will arise which reduce the quality of anomaly images as shown in the third column of Figure \ref{fig::method_saecae}. The first problem is that misalignment between the generated anomalies and masks. Namely, the generated anomaly regions mismatch the specific regions in the masks. On the other hand, the generated anomalies is not pronounced resulting that it looks like the normal images. To circumvent these two issues, we propose  Cross-Attention Enhancement (CAE) in which we re-engineer the cross-attention maps by modifying the vision-text similarity matrix $S_{c}$. CAE improve the similarity between specific visual tokens and text embedding. Furthermore, text embedding contributes more to generation process of visual feature maps and text-guidance in reverse-diffusion process is enhanced. To manipulate the cross-attention maps, mask $M$ is flattened and repeated as $M^c\in \mathbb{R}^{N\times \mathcal{T}}$. We assign enhancement factor in mask $M^c$ as follows:
\begin{equation}
M_{i,j}^c=\left\{\begin{array}{ll}
\alpha, & if \; anomaly\; generation \; at \; location \; (i,j) \\
0, & otherwise,
\end{array}\right.
\end{equation}
where $\alpha$ is CAE factor. The mechanism of SAE is defined as:
\begin{equation}  
S_{c}^e = S_{c} \cdot M_{i,j}^c, \; S_{c}^e\in \mathbb{R}^{N\times \mathcal{T}},
\end{equation}  
\begin{equation}
 A_{c}^e = \text{Softmax}\left(S_{c}^e \right), \;  A_{c}^e \in \mathbb{R}^{N\times \mathcal{T}},
\label{equation4}
\end{equation}
CAE improves alignment between generated anomalies and masks and make generated anomalies more faithful.

\subsection{Self-Attention Enhancement}
To make generated anomalies more realistic and seamless, only considering text guidance and cross-attention maps is not enough. The visual pattern of the whole image as visual context is necessary to be involved in anomaly generation. Guidance by visual context will obviate antipathetic anomalies. We realize visual context guidance in self-attention module, which is also called self-attention enhancement (SAE). Self-attention mechanism is defined as:
\begin{equation}
 Q_{s} = \ell_{q}(\mathcal{F}^l), \;  Q_{s} \in \mathbb{R}^{N\times d_m}, 
\label{equation3}
\end{equation}
\begin{equation}
 K_{s} = \ell_{k}(\mathcal{F}^l), V_{s} = \ell_{v}(\mathcal{F}^l),\;  K_{s},V_{s} \in \mathbb{R}^{N\times d_m},
\label{equation3}
\end{equation}
\begin{equation}  
S_{s} = Q_{s}\left ( K_{s} \right )^{T} /\sqrt{d_{s}}, \; S_{s}\in \mathbb{R}^{N\times N},
\end{equation}  
\begin{equation}
 A_{s} = \text{Softmax}\left(S_{s}\right), \; A_{s} \in \mathbb{R}^{N\times N}.
\label{equation4}
\end{equation}
where $S_{s}$ is vision-vision similarity matrix and $A_{s}$ is self-attention map. The output feature map $O_s$ is defined as $O_s=A_{s}V_{s}, O_s\in \mathbb{R}^{N\times d_m}$. 
In SAE, we improve the the extent to which visual token outside of specific regions in the mask affects the resulting anomalies so as to yield globally coherent anomalies. We also re-weight mask $M$:
\begin{equation}
M_{i,j}^s=\left\{\begin{array}{ll}
0, & if \; anomaly\; generation \; at \; location \; (i,j) \\
\beta, & otherwise,
\end{array}\right.
\end{equation}
where $\beta$ is SAE factor and we use re-weighted mask to regulate self-attention maps as follows:
\begin{equation}  
S_{s}^e = S_{s} \cdot M_{i,j}^s, \; S_{s}^e\in \mathbb{R}^{N\times N},
\end{equation}  
\begin{equation}
 A_{s}^e = \text{Softmax}\left(S_{s}^e \right), \;  A_{s}^e\in \mathbb{R}^{N\times N},
\label{equation4}
\end{equation}
SAE provides visual context for anomaly generation process, which makes generated anomalies blend into original normal image seamlessly.

 \begin{figure}[t]
  \centering
  \centerline{\includegraphics[width=0.82\linewidth]{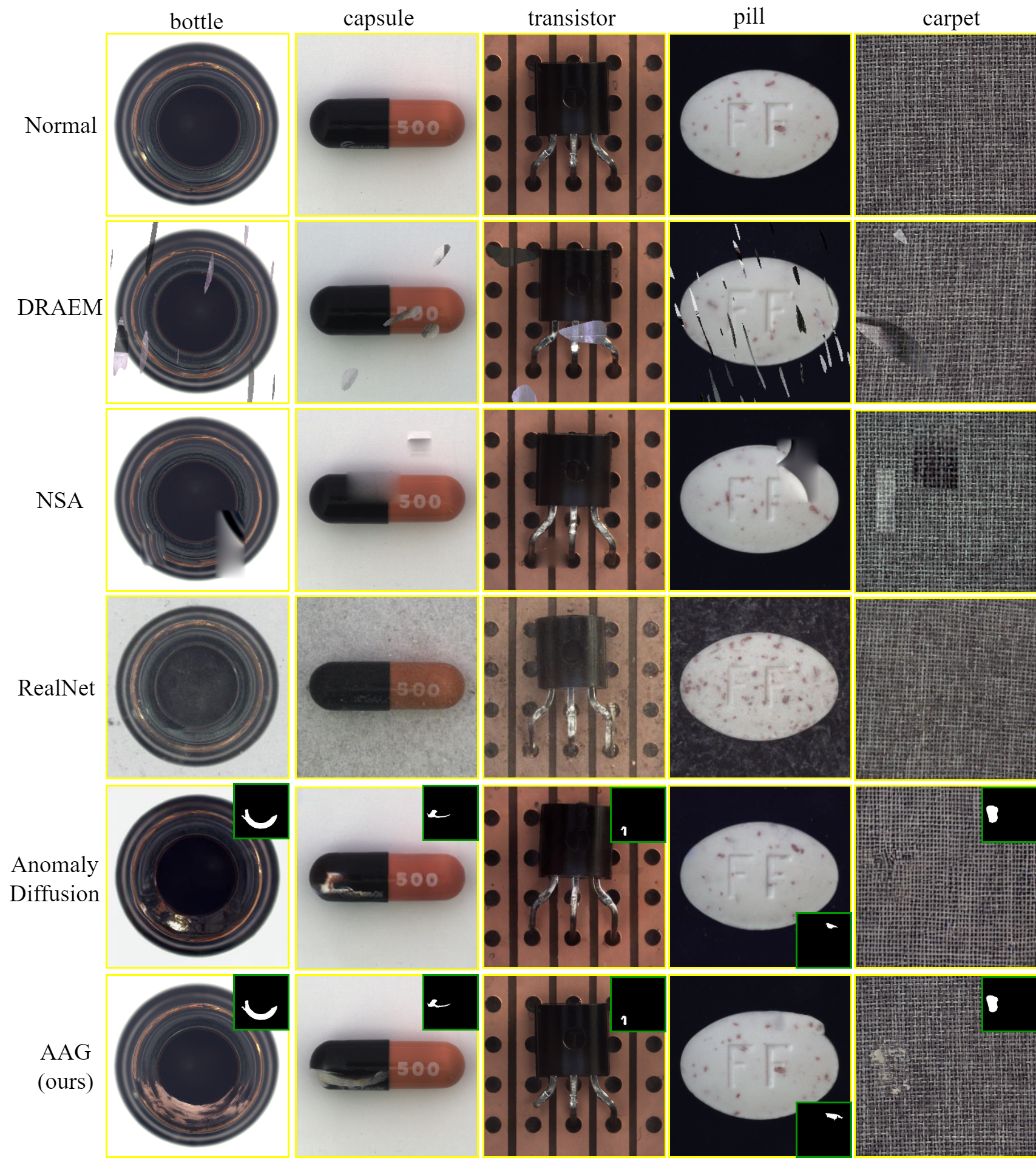}}
  \caption{Comparison in the generation results on
MVTEC AD dataset with other anomaly generation methods.}
  \label{fig::exp_compare_mvtec}
\end{figure}

\subsection{Blended Mechanism}
To generate anomalies in specific regions according to mask and preserve other regions in origin normal image, we utilize blended mechanism inspired by Attentive Eraser~\cite{ae} in anomaly generation process. Let $z_0$ denotes visual embedding and $z_t$ denotes noised visual embedding at timestep $t$ diffused as Eq.(\ref{eq:xt}). At each generation step, U-net model with SAE and CAE $\epsilon_{\theta}\left(x_{t}, \mathcal{P}_e, t\right)$ is used to predict noise and we sample noisy visual embedding  $z_{t-1}^{init}$: 
\begin{equation}
\begin{aligned}
z_{t-1}^{init}=\frac{1}{\sqrt{\alpha_{t}}}\left(z_{t}-\frac{1-\alpha_{t}}{\sqrt{1-\bar{\alpha}_{t}}} \epsilon_{\theta}\left(x_{t}, \mathcal{P}_e, t\right)\right)+\sigma_{t} n,
\end{aligned}
\label{eq:zt-1init}
\end{equation}
where $n \sim \mathcal{N}(\mathbf{0}, \mathbf{I})$. To perform blended mechanism in generation process, we obtain noisy visual embedding $z_{t}^0$ at timestep $t$ from initial $z$ following Eq.(\ref{eq:xt}).
Then we blend $z_{t}^0$ and $z_{t-1}$ based on mask $M$:
\begin{equation}
z_{t-1} = z_{t-1}^{init} \cdot M + z_{t}^0  \cdot (1-M) 
\label{eq:zt-1}
\end{equation}
where $z_{t-1}$ contains original visual pattern and generated anomalies pattern. Blended mechanism can avert to damage normal regions in some regions where it is unexpected to generate anomalies.

Each step  of anomaly generation in AAG framework is summarized in Algorithm 1.

 \begin{figure}[t]
  \centering
  \centerline{\includegraphics[width=0.88\linewidth]{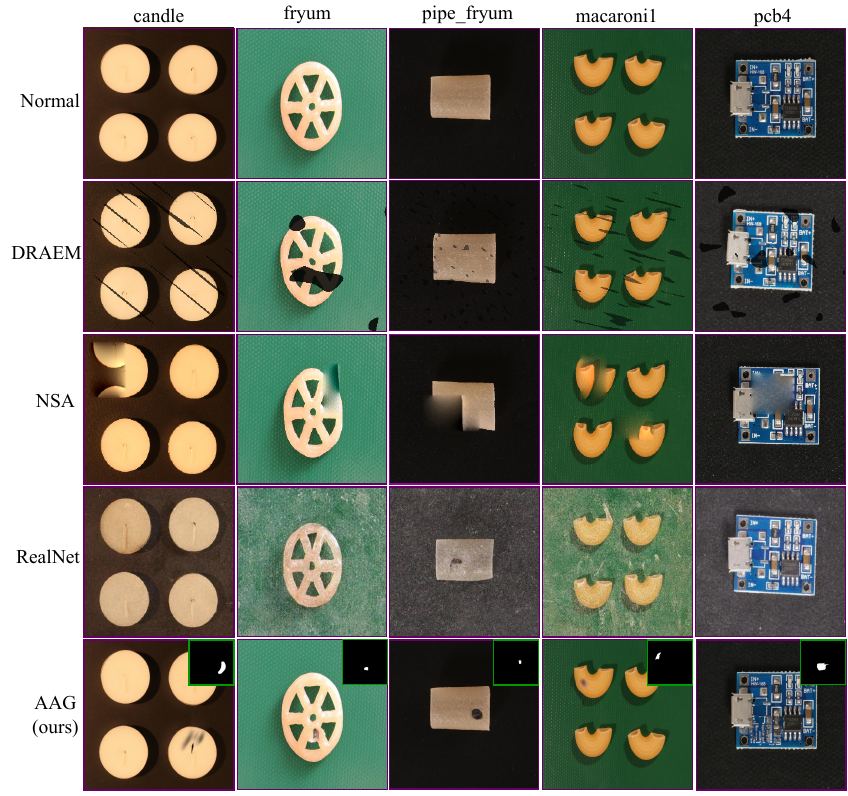}}
  \caption{Comparison in the generation results on
VisA dataset with other anomaly generation methods.}
  \label{fig::exp_compare_visa}
\end{figure}

\section{Experiments}
\subsection{Experimental Setup}
\subsubsection{Dataset.}
We conduct generation experiments on MVTEC AD~\cite{mvtec} and VisA~\cite{visa} datasets. There are totally 27 kinds of object classes in MVTEC AD and VisA datasets.
\subsubsection{Metric.}
For anomaly generation, we use Inception Score (IS)~\cite{is} and the Intra-cluster Pairwise Learned Perceptual Image Patch Similarity (IC-LPIPS) to evaluate the generation quality and diversity. To evaluate the effectiveness of our method in downstream anomaly detection tasks, we utilize Under the
Receiver Operating Characteristic (AUROC), Area Under the Precision-Recall curve (AUPR) and Per-Region-Overlap (PRO) metrics.

\begin{table}[t]
\setlength{\tabcolsep}{5pt}
  \centering
  \resizebox{0.9\linewidth}{!}{
    \begin{tabular}{c|llll}
    \toprule
      &  \multicolumn{2}{c}{Detection} & \multicolumn{2}{c}{Localization} \\
          Method & I-AUROC & I-AUPR & P-AUROC & PRO \\
    \midrule
     MemSeg+RandomDisturb   & 83.6   & 93.0  & 77.6  & 79.4   \\
           MemSeg+NSA   & 91.0   & 95.6  & 84.1  & 79.8   \\
          MemSeg+DRAEM    & 92.4  & 96.1  & 85.5  & 84.8    \\
    MemSeg+AnomalyDiffusion     & 91.4   & 96.7  & 86.7  & 83.2   \\
    MemSeg+AAG     & \textbf{93.4}   & \textbf{97.4}  & \textbf{88.1}  & \textbf{85.6}   \\
    \midrule
    RRD+RandomDisturb   & 98.6   & 99.4  & 98.0  & 94.7   \\
     RRD+NSA   & 98.1   & 99.1 & 97.7  & 94.2   \\
          RRD+DRAEM    & 98.2  & 99.4  & 97.9  & 94.0    \\
    RRD+AnomalyDiffusion     & 98.6   & 99.4  & 97.8  & 94.2   \\
    RRD+AAG     & \textbf{98.9}   & \textbf{99.6}  & \textbf{98.1}  & \textbf{94.8}   \\
    \bottomrule
    \end{tabular}}%
    \caption{Anomaly detection performance comparison of AAG against other anomaly generation methods on MVTEC-AD dataset. The best results are highlight in bold.}
  \label{tab:edt::mvtec}%
\end{table}%

\subsubsection{Implementation Details.}
Our proposed AAG is based on pretrained Stable Diffusion XL 1.0 model~\cite{sdxl}. In our method, we use masks generated by Anomalydiffusion~\cite{anomalydiffusion}. Timestep $T$ is set to 50. The text prompt template is "A {cls} that is damaged and broken" where cls refers to object classes. It is not necessary to design meticulously text prompt and obtain description of anomalies in testing samples. CAE factor $\alpha$ is set to 1.5 and SAE factor $\beta$ is set to 1.1 experimentally. We conduct all experiments on NVIDIA RTX 4090 GPU.

\begin{table}[ht]
\setlength{\tabcolsep}{5pt}
  \centering
  \resizebox{0.9\linewidth}{!}{
    \begin{tabular}{c|llll}
    \toprule
      &  \multicolumn{2}{c}{Detection} & \multicolumn{2}{c}{Localization} \\
          Method & I-AUROC & I-AUPR & P-AUROC & PRO \\
    \midrule
     MemSeg+RandomDisturb   & 73.0   & 77.4  & 67.2  & 66.7   \\
           MemSeg+NSA   & 80.7   & 85.9  & 87.4  & 81.1   \\
          MemSeg+DRAEM    & 83.4  & 86.7  & 76.7  & 76.9    \\
    MemSeg+AAG     & \textbf{86.5}   & \textbf{89.1}  & \textbf{87.6}  & \textbf{89.0}   \\
    \midrule
    RRD+RandomDisturb   & 91.7   & 92.0  & 98.0  & 90.8   \\
     RRD+NSA   & 91.6   & 92.1  & 98.3  & 91.9   \\
          RRD+DRAEM    & 91.1  & 91.6  &97.9  & 90.6    \\
    RRD+AAG     & \textbf{96.2}  & \textbf{96.6}  & \textbf{98.4}  & \textbf{92.9}   \\
    \bottomrule
    \end{tabular}}%
    \caption{Anomaly detection performance comparison of AAG against other anomaly generation methods on VisA dataset.}
  \label{tab:edt::visa}%
\end{table}%

 \begin{figure}[t]
  \centering
  \centerline{\includegraphics[width=0.88\linewidth]{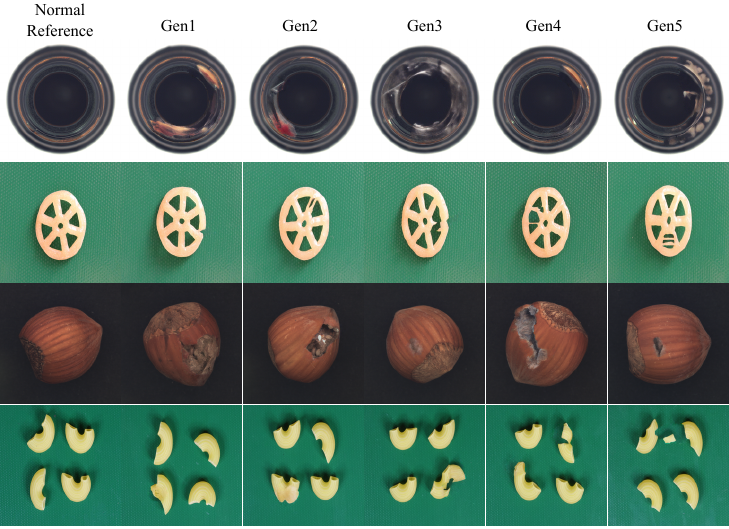}}
  \caption{Illustration of generation diversity of AAG. We present 5 generation examples for each category.}
  \label{fig::exp_diversity}
\end{figure}

\subsection{Comparison in Anomaly Generation}
\subsubsection{Anomaly generation quality}
To quantitatively measure the anomaly generation quality, we use Inception Score (IS) metric as an indicator. As shown in Table \ref{tab::isil}, our proposed AAG achieves the state-of-the-art IS of 1.94, which is 0.1$\uparrow$ higher than the previous SOTA AnomalyControl~\cite{anomalycontrol} and is 0.14$\uparrow$ higher than AnomalyDiffusion~\cite{anomalydiffusion}. AAG achieves the highest IS on 10 classes on MVTEC AD, which accounts for about 67\% of the total classes. It is notable that AAG improves the IS by a significant 1.1$\uparrow$ on zipper class and 0.45$\uparrow$ on carpet class compared with the second-best results. The high generation quality of AAG is illustrated by IS metric. However, IS indicator is not intuitive and fails to demonstrate what anomaly images AAG generates. We visualize the synthesized anomaly images generated by AAG and other methods.

The generated anomaly images on MVTEC AD and VisA datasets are shown in Figure \ref{fig::exp_compare_mvtec} and Figure \ref{fig::exp_compare_visa}. It is demonstrated that anomaly images generated by AAG are more realistic and high-fidelity than other methods. DRAEM~\cite{draem} and NSA~\cite{nsa} use cut-paste and Poisson image editing strategy. They generate anomalies roughly, which is not coherent with original normal images and introduces some unusual patterns. Anomaly images generated by RealNet~\cite{realnet} are noisy and anomalies are not obvious. Though AnomalyDiffusion is trained using anomalous data in test set of MVTEC AD dataset, the anomaly images generated by AnomalyDiffusion are still less realistic compared to AAG shown in Figure \ref{fig::exp_compare_mvtec}. Additionally, we showcase the generation results on VisA dataset in Figure \ref{fig::exp_compare_visa}, which further demonstrates superior generation quality of AAG. Notably, AAG presents globally coherent anomaly generation even on complicated PCB images. We present synthesized anomaly images on all 27 classes in MVTEC AD and VisA datasets in supplementary material.

\subsubsection{Anomaly generation diversity}
In addition to generation quality, we also evaluate the generation diversity of AAG. As shown in Table \ref{tab::isil}, our AAG achieves the state-of-the-art average IL of 0.35, which illustrates the generalization and diversity of synthetic anomalies generated by AAG. We showcase 5 different kinds of synthetic anomalies for each category generated by AAG in Figure \ref{fig::exp_diversity}. It is demonstrated that AAG can generate realistic anomalies with diverse appearances and patterns.

\subsection{Ablation Study}

\subsubsection{Effectiveness in Downstream tasks}
To evaluate the effectiveness of AAG and its contribution to downstream anomaly detection tasks. We apply AAG in some anomaly detection methods in which anomaly generation is used for training. For comparisons, we select MemSeg~\cite{memseg} and RRD~\cite{rd++} as anomaly detection methods and compare AAG with other anomaly generation methods: RandomGuass~\cite{cdo}, NSA~\cite{nsa}, DRAEM~\cite{draem} and AnomalyDiffusion~\cite{anomalydiffusion}. As illustrated in Table \ref{tab:edt::mvtec} and Table \ref{tab:edt::visa}, our AAG shows
higher performance improvement in anomaly detection than other anomaly generation methods.  For example, AAG achieves (2.0\%, 2.4\%) and (0.3\%, 0.6\%) improvements in terms of (I-AUROC, PRO) in MemSeg and RRD respectively compared to AnomalyDiffusion~\cite{anomalydiffusion} on MVTEC AD dataset. As for VisA dataset, AAG also achieves the highest performance improvement compared with previous anomaly generation methods showcased in Table \ref{tab:edt::visa}. Notably, AAG improves I-AUROC by a significant 3.1\% and 5.1\% on VisA dataset.

\begin{table}[htbp]
  \centering
  \caption{Ablation Studies on CAE and SAE.}
  \resizebox{0.85\linewidth}{!}{
    \begin{tabular}{cccccc}
    \hline
    \multirow{2}[0]{*}{CAE} & \multirow{2}[0]{*}{SAE} & \multicolumn{2}{c}{MVTec} & \multicolumn{2}{c}{VisA} \\
          &       & I-AUROC & PRO   & I-AUROC & PRO    \\
          \hline
    \usym{2717}     & \usym{2717}     & 98.5  & 94.0    & 93.3  & 91.6   \\
    \usym{2714}     &\usym{2717}     & 98.7  & 94.1  & 93.7  & 92.1   \\
    \usym{2714}   & \usym{2714}     & \textbf{98.9}     &  \textbf{94.7}  &    \textbf{96.3}   &   \textbf{93.0}   \\
    \hline
    \end{tabular}}
  \label{abla::caesae}%
\end{table}%

\subsubsection{Effectiveness of Cross-Attention Enhancement}
We design Cross-Attention Enhancement (CAE) module to enhance the guidance of text prompt, which avoids failing to generate anomalies. To validate the the effectiveness of CAE, we use synthetic anomalies without CAE and SAE to train RRD. Table \ref{abla::caesae} presents the comparison results between with and without CAE and SAE. The anomaly detection performance without CAE and SAE drops by 0.2\% and 0.4\% in terms of I-AUROC on MVTEC AD and VisA datasets compared to performance with CAE. As for anomaly localization performance, CAE achieves 0.1\% and 0.5\% improvements in PRO score. We visualize the synthetic anomalies without CAE and SAE for further demonstration in Figure \ref{fig::method_saecae}. As shown in Figure \ref{fig::method_saecae}, synthetic anomalies are not pronounced or not natural without CAE and SAE.

\subsubsection{Effectiveness of Self-Attention Enhancement}
We also validate the effectiveness of SAE in downstream anomaly detection tasks based on RRD. As shown in Table \ref{abla::caesae}, the anomaly detection performance is improved by 0.2\%$\uparrow$ and 2.6\%$\uparrow$ in terms of I-AUROC and anomaly localization performance is improved by 0.6\%$\uparrow$ and 0.9\%$\uparrow$ in PRO score when SAE is introduced. Figure \ref{fig::method_saecae} shows the synthetic anomalies only with CAE and high-quality synthetic anomalies generated by AAG. Incorporating SAE, generated anomalies are more coherent and realistic compared with generation only with CAE.

 \begin{figure}[t]
  \centering
  \centerline{\includegraphics[width=0.84\linewidth]{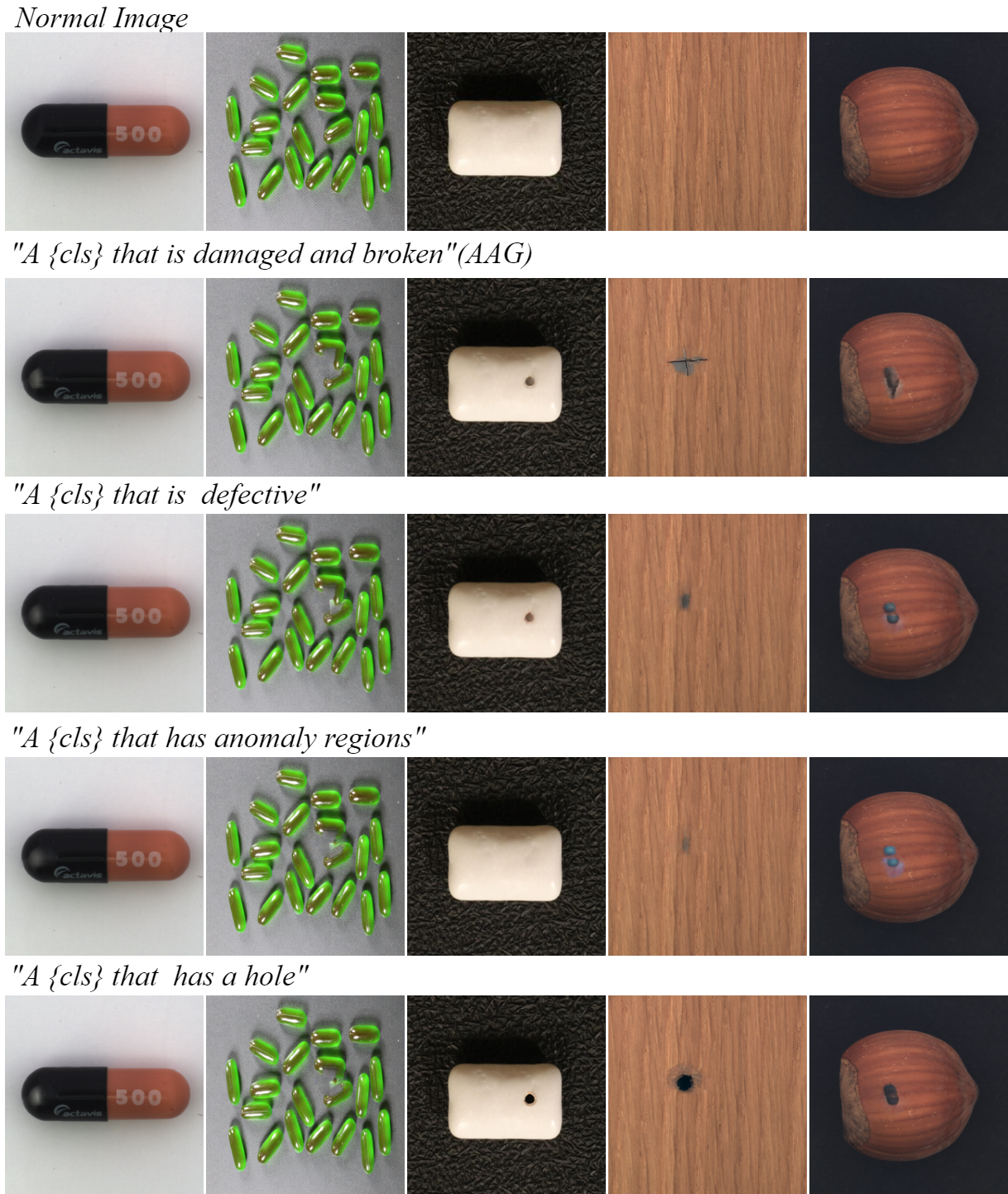}}
  \caption{Generation results in different text prompts.}
  \label{fig::prompt}
\end{figure}

\subsubsection{Ablations on different text prompts}
AAG is an anomaly type-agnostic generation method. Text prompt used in AAG only includes object category without descriptions about anomalies . Figure \ref{fig::prompt} presents synthetic anomalies under different text prompts. The \textit{\{cls\}} represents the category of object. It is demonstrated that AAG can generate high-quality anomalies with no matter general text prompt (e.g., "\textit{A \{cls\} that is anomalous}") or pattern-specific text prompt (e.g., "\textit{A \{cls\} that has a hole}"). Therefore, it is not necessary for users to design text prompts meticulously. Meanwhile, users can customize some pattern-specific text prompts for guidance as shown in the fifth row of Figure \ref{fig::prompt}.

\section{Conclusion}
In this paper, we propose a simple but effective framework dubbed AAG for anomaly generation, which dramatically simplifies the anomaly generation framework and pushes the limits of generation quality and diversity. Our proposed AAG incorporates Cross-Attention Enhancement (CAE) to enhance the guidance of text prompts for pronounced anomalies generation. Meanwhile, we design Self-Attention Enhancement (SAE) to improve global coherence and fidelity of generated anomalies. AAG needs no anomaly data for training and types of anomalies for captions. AAG undergoes rigorous evaluation on two datasets and outperforms state-of-the-art methods in anomaly generation. Notably, AAG exhibits notable performance improvement in downstream anomaly detection tasks.

\bibliography{reference}

\newpage
\newpage
\section{Appendix}
\begin{figure*}[ht]
  \centering
  \centerline{\includegraphics[width=0.85\linewidth]{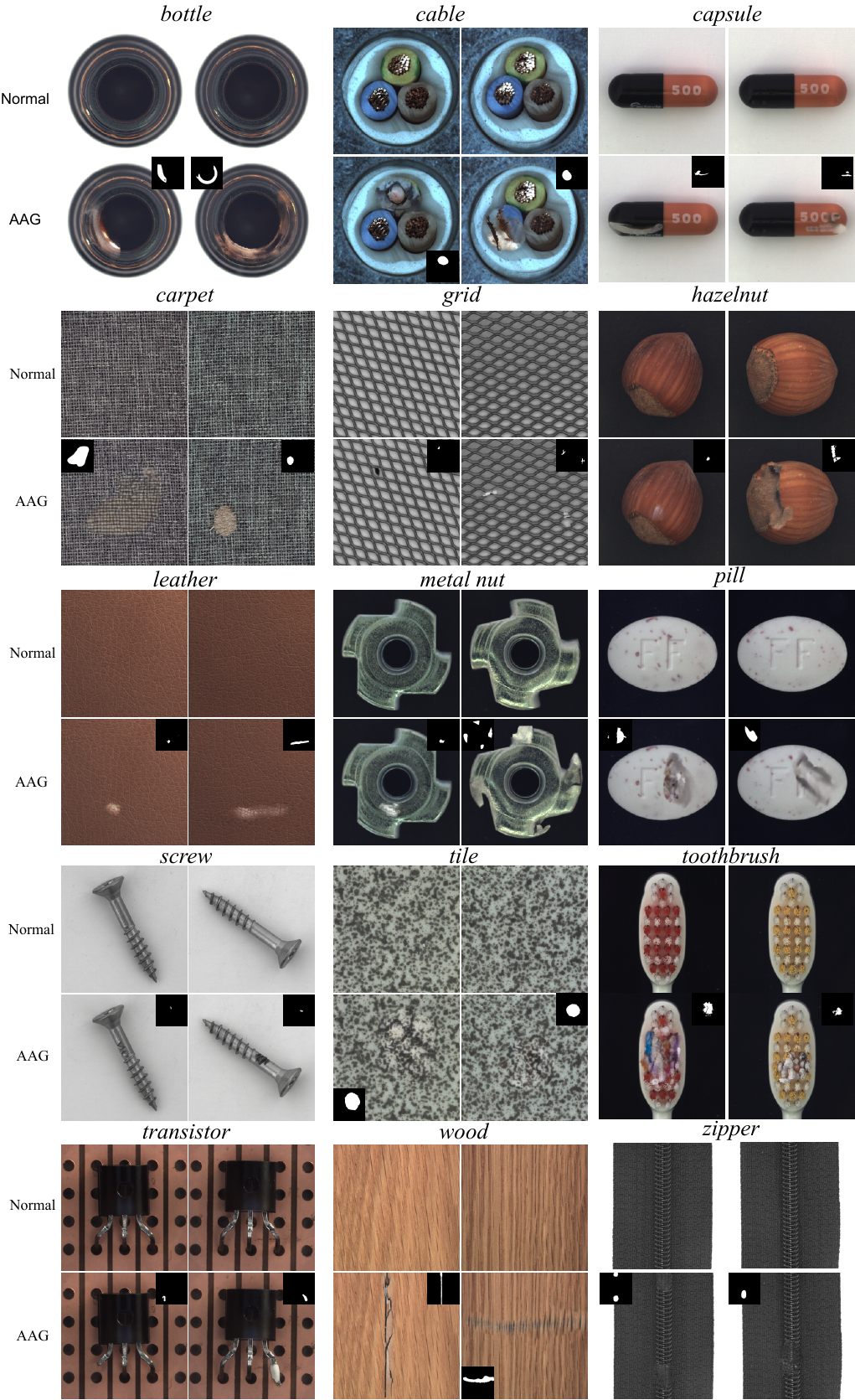}}
  \caption{Generation results on all categories of MVTEC AD dataset.}
  \label{fig::allmvtec}
\end{figure*}

\begin{figure*}[ht]
  \centering
  \centerline{\includegraphics[width=0.85\linewidth]{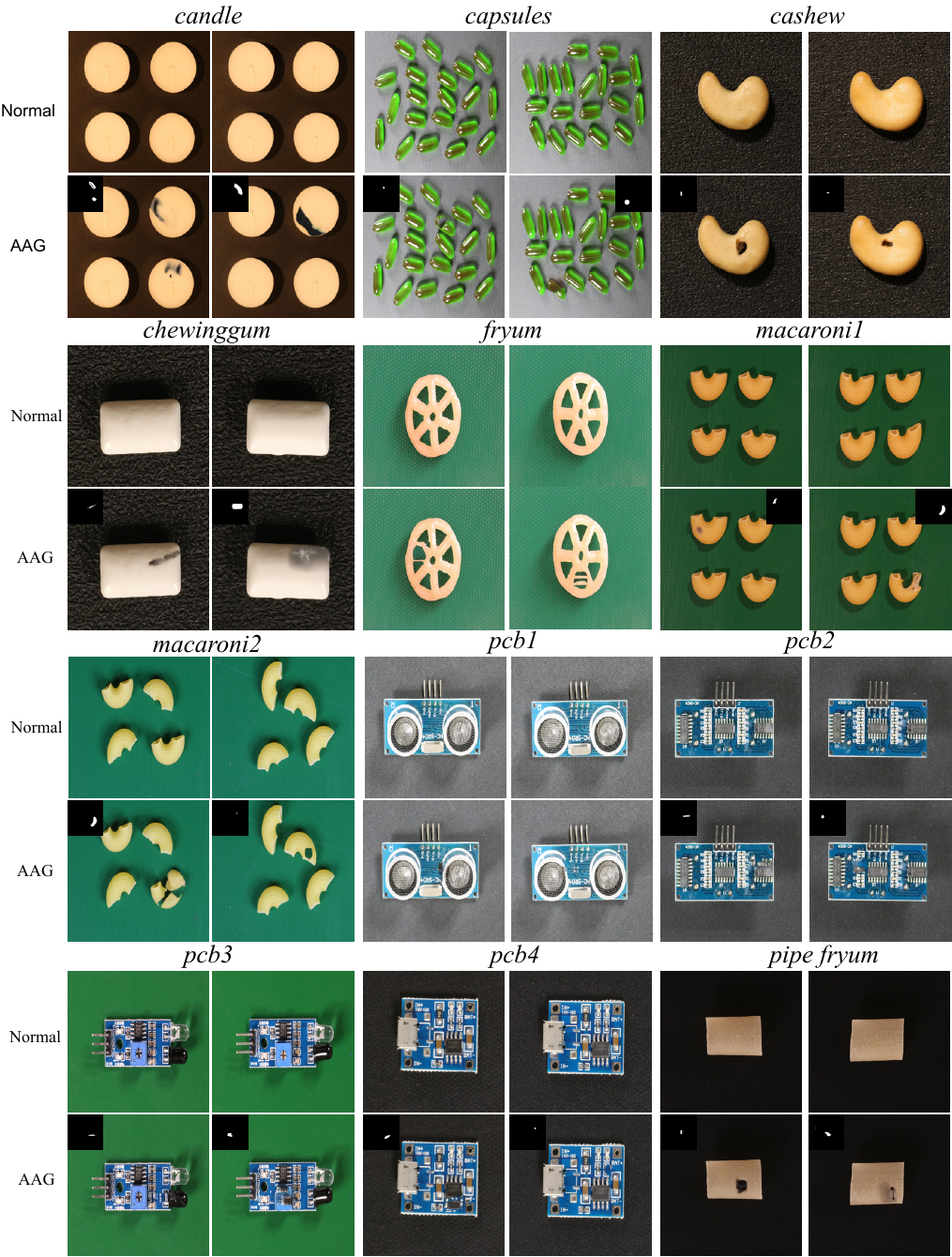}}
  \caption{Generation results on all categories of VisA dataset.}
  \label{fig::allvisa}
\end{figure*}

We present synthesized anomaly images on all 27 categories of MVTEC AD and VisA datasets in Figure \ref{fig::allmvtec} and Figure \ref{fig::allvisa}. We provide two anomaly generation examples for each category.

\end{document}